\pdfoutput=1 
\documentclass[11pt]{article}
\usepackage[final]{coling}
\usepackage{times}
\usepackage{latexsym}
\usepackage[T1]{fontenc}
\usepackage[utf8]{inputenc}
\usepackage{amsmath}
\usepackage{amssymb}
\usepackage{microtype}
\usepackage{inconsolata}
\usepackage{graphicx}
\usepackage{float} 
\usepackage{amssymb}
\usepackage{multirow}
\usepackage{booktabs} 
\title{DualKanbaFormer: An Efficient Selective Sparse Framework for Multimodal Aspect-based Sentiment Analysis
}

\author{
	Adamu Lawan\textsuperscript{1}, Juhua Pu\textsuperscript{1}\thanks{Corresponding author}, Haruna Yunusa\textsuperscript{2}, Aliyu Umar\textsuperscript{3}, Muhammad Lawan\textsuperscript{4}, \\
    \textbf{Adamu Sani Yahaya, Mahmoud Basi }\\
	\textsuperscript{1}School of Computer Science and Technology, Beihang University, Beijing, China \\
	\textsuperscript{2}School of Automation Science and Electrical Engineering, Beihang University, Beijing, China \\
	\textsuperscript{3}School of Computing, University of Portsmouth, Portsmouth, UK \\
	\textsuperscript{4}Department of Information and Communication Technology, Federal University, Gusau, Nigeria \\
	\texttt{\{alawan, pujh, yunusa2k2\}@buaa.edu.cn}
}

\begin{document}
	\maketitle

	\begin{abstract}
	Multimodal Aspect-based Sentiment Analysis (MABSA) enhances sentiment detection by integrating textual data with complementary modalities, such as images, to provide a more refined and comprehensive understanding of sentiment.  However, conventional attention mechanisms, despite notable benchmarks, are hindered by quadratic complexity, limiting their ability to fully capture global contextual dependencies and rich semantic information in both modalities. To address this limitation, we introduce DualKanbaFormer, a novel framework that leverages parallel Textual and Visual KanbaFormer modules for robust multimodal analysis. Our approach incorporates Aspect-Driven Sparse Attention (ADSA) to dynamically balance coarse-grained aggregation and fine-grained selection for aspect-focused precision, ensuring the preservation of both global context awareness and local precision in textual and visual representations. Additionally, we utilize the Selective State Space Model (Mamba) to capture extensive global semantic information across both modalities. Furthermore, We replace traditional feed-forward networks and normalization with Kolmogorov-Arnold Networks (KANs) and Dynamic Tanh (DyT) to enhance non-linear expressivity and inference stability. To facilitate the effective integration of textual and visual features, we design a multimodal gated fusion layer that dynamically optimizes inter-modality interactions, significantly enhancing the model’s efficacy in MABSA tasks. Comprehensive experiments on two publicly available datasets reveal that DualKanbaFormer consistently outperforms several state-of-the-art (SOTA) models.
\end{abstract}

\section{Introduction}
MABSA is a growing field that aims to determine sentiment polarity for specific aspects within multimodal data, such as text and images. This fine-grained analysis is crucial for applications like product marketing and customer feedback analysis, where understanding sentiment toward particular features is Important. \cite{Xu2019, Yu2019} pioneered the combination of text and visual data for more accurate aspect-based sentiment evaluation. Recent advancements include the Entity-Sensitive Attention and Fusion Network (ESAFN) by \cite{Yu2020}, the Knowledge-enhanced Framework (KEF) by \cite{Zhao2022}, and the Image-Target Matching network (ITM) by \cite{Yu2022}, which uses various attention mechanisms to fuse the representation from image and text, improving sentiment analysis accuracy significantly. 
The Vision Language Pre-training framework (VLP-MABSA) by \cite{Ling2022} uses a single architecture to integrate language, vision, and multimodal pre-training tasks, enhancing the identification of fine-grained aspects and opinions across text and images. Further innovations such as CoolNet \cite{Xiao2022} and the Aspect-Guided Multi-View Interactions and Fusion Network (AMIFN) by \cite{Yang2024c} focus on better text-image interactions by leveraging syntactic information. The Hierarchical Interactive Multimodal Transformer (HIMT) by \cite{Yu2023} and the Multi-level Attention Map Network (MAMN) by \cite{Xue2023} offer advanced methods for reducing semantic gaps and filtering noise in MABSA. Most recently, the Multi-Grained Fusion Network with Self-Distillation (MGFN-SD) by \cite{Yang2024b} and the Global–Local Features Fusion with Co-Attention (GLFFCA) by \cite{Wang2024} have introduced techniques to integrate coarse- and fine-grained features and capture multimodal associations, enhancing MABSA accuracy.

Despite achieving significant performance benchmarks, attention mechanisms continue to face considerable challenges in modeling global context dependencies effectively due to their quadratic computational complexity. This constraint limits their ability to fully leverage the rich semantic information inherent in textual and visual data, thereby restricting their capacity to provide a comprehensive, context-aware understanding of both modalities. Addressing these limitations is critical to improving performance and unlocking new possibilities in applications like MABSA, where precise and nuanced interpretation of text and images is paramount.

Drawing inspiration from the Native Sparse Attention (NSA) architecture proposed by Yuan et al. \cite{yuan2025native}, renowned for its efficiency in processing long-range sequences and capturing global contextual dependencies through a sparse attention mechanism, we introduce DualKanbaFormer, an innovative framework featuring parallel Textual KanbaFormer and Visual KanbaFormer modules. These modules are specifically designed to process textual and visual data, respectively, with a focus on aspect-specific features critical for MABSA. Each KanbaFormer integrates five key components to achieve robust intra-modal representation:
(1) Mamba, a selective state space model that captures the global semantic richness of visual data and models long-range dependencies between aspects and opinion words in text, enabling comprehensive context understanding across modalities.
(2) Aspect-Driven Sparse Attention (ADSA), this dynamic hierarchical sparse strategy combines coarse-grained token aggregation with fine-grained token selection. ADSA ensures the retention of global context awareness and local precision in both textual and visual representations.
(3) Kolmogorov-Arnold Networks (KANs), KANs model complex non-linear patterns for enhanced representation.
(4) Dynamic Tanh (DyT), motivated by the observation that layer normalization in DualKanbaFormer often produces tanh-like, S-shaped input-output mappings, we replace it with DyT. This substitution eliminates normalization while improving flexibility, with experiments showing that DualKanbaFormers without normalization consistently outperform their normalized counterparts, typically requiring minimal hyperparameter tuning.
(5) Intra-Modal Gated Fusion, this mechanism integrates insights from Mamba and the KanFormer substructure, our term for the ADSA and KAN combination, a transformer variant with superior non-linear modeling capabilities. 
To facilitate dynamic inter-modality interaction, we introduce a multimodal gated fusion layer that unifies the outputs of the Textual KanbaFormer and Visual KanbaFormer. This layer enhances cross-modal synergy, significantly boosting the model’s effectiveness for MABSA tasks by enabling precise and nuanced interpretation of aspect-specific textual and visual data.
Contributions:
\begin{itemize}
\item We introduce DualKanbaFormer, a novel architecture that integrates parallel Textual KanbaFormer and Visual KanbaFormer modules to synergistically model global semantic richness in visual and textual data. By leveraging Mamba and Aspect-Driven Sparse Attention (ADSA), this framework achieves efficient and comprehensive cross-modal context understanding, advancing aspect-specific sentiment analysis.
\item To the best of our knowledge, this work represents the first application of Selective State Space Models (SSMs) and sparse attention mechanisms to capture the global semantic richness of images within MABSA. This pioneering approach establishes a new paradigm for multimodal sentiment understanding, bridging efficient sequence modeling with aspect-focused visual analysis.
\item Extensive experiments on two benchmark datasets for MABSA reveal that DualKanbaFormer consistently outperforms some SOTA methods, showcasing its superior performance.
\end{itemize}

\section{Related Work}
Aspect-based Sentiment Analysis (ABSA) has significantly improved by integrating attention mechanisms and pre-trained models. Attention-
based Long Short-term Memory network (AE-LSTM) \cite{Wang2016} pioneered the use of aspect information to identify key sentence components via attention. Interactive Attention Networks (IAN) \cite{Ma2017} further enhanced this by modelling aspect-sentence interactions for sentiment analysis. Multi-grained Attention Network (MGAN) \cite{Fan2018} introduced a multi-grained attention network, refining the integration of aspects and sentences from coarse to fine granularity. To capture nuanced relationships between aspects and sentences, Bidirectional Encoder Representations from Transformers (BERT) \cite{Devlin2018} leveraged pre-trained representations and attention mechanisms, setting a new benchmark in ABSA performance. These approaches collectively highlight the evolution of ABSA towards more sophisticated and context-aware sentiment analysis.

MABSA has also progressed significantly, driven by the need to assess sentiment towards specific aspects of multimodal content. Early work by \cite{Xu2019} and \cite{Yu2019} introduced foundational models like the Multi-Interactive Memory Network (MIMN) and Target-oriented Multimodal Sentiment Classification (TMSC), which integrated text and visual data to enhance sentiment accuracy through multi-modal feature fusion. Subsequent research focused on refining attention mechanisms and cross-modal alignment. \cite{Yu2020} and \cite{Zhao2022} improved sentiment analysis by introducing entity-sensitive attention networks and knowledge-enhanced frameworks, which better aligned and fused text-image representations, increasing precision in sentiment prediction.
    
Recent developments emphasized pre-training, knowledge transfer, and multi-grained representation learning. \cite{Ling2022} and \cite{Yu2022} introduced VLP-MABSA and ITM, integrating multimodal pre-training tasks and refining image-target matching for enhanced cross-modal alignment. \cite{Xiao2023, Yang2024c} further improved ABSA by focusing on better text-image interactions and leveraging syntactic information. Innovative approaches like MGFN-SD \cite{Yang2024b} and GLFFCA \cite{Wang2024} combined coarse- and fine-grained feature integration with advanced fusion techniques, achieving higher accuracy and robustness in sentiment prediction.

Another notable area of research involves combining the Transformer and state space model (SSM)-based models \cite{gu2020, gu2021, gu2023} for language modeling \cite{Fathi2023, Lieber2024, Park2024, Xu2024} and image processing \cite{Dosovitskiy2020, Zhu2024}. \cite{lawan-etal-2025-enhancing} integrated Transformer and Mamba to capture long-range dependency in Aspect-based Sentiment Analysis (ABSA). \cite{Li_Liu_Li_Wang_Liu_Liu_Chen_Yuan_2025} used KANs \cite{liu2024} for medical image segmentation and generation.
The evolution of MABSA methodologies has prioritized advancements in multimodal feature fusion, attention mechanisms, and cross-modal alignment, driving significant progress in the field's capabilities.

\begin{figure*}[t] 
	\centering
	\includegraphics[width=\textwidth]{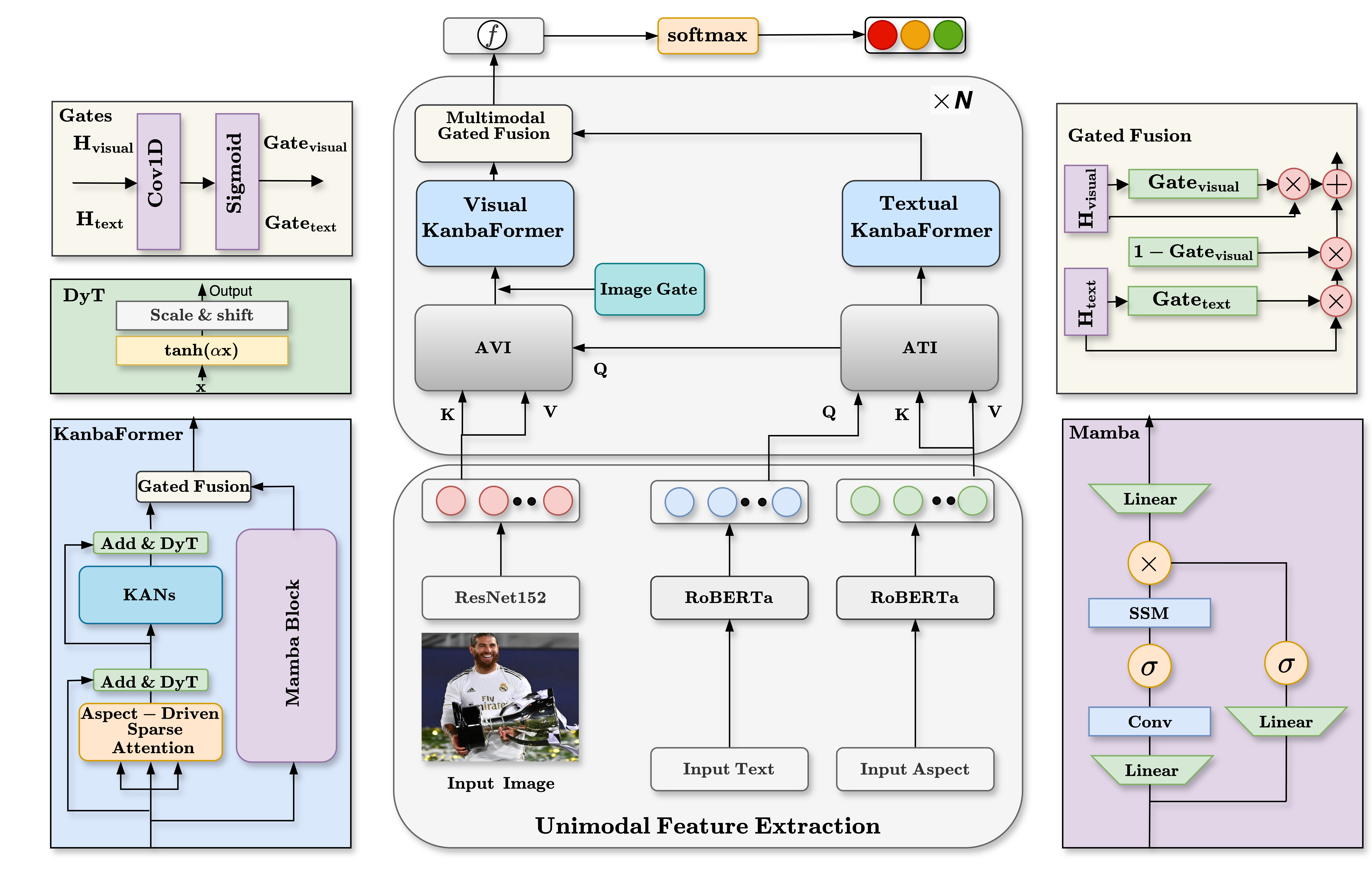}
	\caption{DualKanbaFormer complete architecture}
	\label{fig:DualKanbaFormer}
\end{figure*}

\section{Proposed DualKanbaFormer Model}
Figure~\ref{fig:DualKanbaFormer} depicts the DualKanbaFormer architecture for MABSA. The framework integrates textual and visual data through four components: the Unimodal Feature Extraction Module, the Multi-Interactive Attention Module, the DualKanbaFormer Module and Multimodal Gated Fusion Module. Subsequent subsections detail each component’s role.
\subsection{Task Formulation}

In multimodal sentiment analysis, each sample, such as a user post or review, consists of a sentence \( S \) containing \( n \) words (\( w_1, w_2, \dots, w_n \)), an associated image \( V \), and \( r \) aspect terms (\( A_1, A_2, \dots, A_r \)). Aspect terms \( A_i \) where \( 1 \leq i \leq r \), may either explicitly appear in the sentence or be implied. Given a sample comprising a sentence-image pair (\( S, V \)) and an aspect \( A \), the task of MABSA is to determine the sentiment polarity \( y \) related to the given aspect \( A \), where \( y \) can be \textit{positive}, \textit{negative}, or \textit{neutral}.

\subsection{Unimodal Feature Extraction}

RoBERTa \cite{Liu2019} encoders generate hidden representations for aspect terms and sentences, with aspect terms framed using special tokens and encoded into aspect representations \( \mathbf{H}_A \). Similarly, sentences are encoded to produce context-sensitive representations \( \mathbf{H}_S \). Visual features are obtained from the final convolutional layer of a pre-trained 152-layer ResNet model \cite{He2016}, where images are resized and processed to extract features, as described below:

\begin{equation}
    \mathbf{H}_I = \text{ResNet}_{152}(\tilde{V})
\end{equation}
The extracted image feature, denoted as \( \mathbf{H}_I \in \mathbb{R}^{2048 \times 7 \times 7} \), consists of 2048-dimensional visual features for each of the \( 7 \times 7 \) visual blocks.

\subsection{Multi-Interactive Attention Module}
\subsubsection{Aspect-Textual Interaction (ATI)}

To extract important textual semantic information related to an aspect, we use the Aspect-Textual Interaction (ATI) module, as shown in Fig.~\ref{fig:DualKanbaFormer}. ATI includes fully connected layers and multi-head cross-modal attention (MHCA) \cite{Tsai2019} to enhance textual representations by focusing on the aspect. 
\begin{equation}
    \tilde{\mathbf{H}}_{A \to S} = \text{MHCA}(\mathbf{H}_S, \mathbf{H}_A, \mathbf{H}_A)
\end{equation}
\begin{equation}
    \mathbf{H}_{AS} = \text{FC}(\tilde{\mathbf{H}}_{A \to S}) + \tilde{\mathbf{H}}_{A \to S}
\end{equation}

\subsubsection{Aspect-Visual Interaction (AVI)}
Previous studies have demonstrated the advantages of multimodal fusion in MABSA, highlighting the necessity of obtaining high-quality visual representations to enhance model performance and robustness. We employ a MHCA mechanism to extract visual information pertinent to the task, as shown in Fig.~\ref{fig:DualKanbaFormer}. The process is described below:
\begin{equation}
    \tilde{\mathbf{H}}_{AS \to I} = \text{MHCA}(\mathbf{H}_{AS}, \tilde{\mathbf{H}}_I, \tilde{\mathbf{H}}_I)
\end{equation}
\begin{equation}
    \mathbf{H}_{AS \to I} = \text{FC}(\tilde{\mathbf{H}}_{AS \to I}) + \tilde{\mathbf{H}}_{AS \to I}
\end{equation}

The fine-grained visual representation \( \mathbf{H}_{AS \to I} \) can introduce noise due to irrelevant aspects in the images and sentences with low relevance \cite{Yu2022}. To mitigate this, we apply max-pooling to extract the most significant visual features \( \mathbf{h}_{AS \to I} \in \mathbb{R}^{d \times 1} \) and use a gating mechanism to filter out noise:

\begin{equation}
    \mathbf{h}_{AS \to I} = \text{maxpooling}(\mathbf{H}_{AS \to I})
\end{equation}
\begin{equation}
    \mathbf{g}_I = \sigma(\mathbf{W}_G \mathbf{h}_{AS \to I} + \mathbf{b}_G)
\end{equation}
where \( \mathbf{W}_G \in \mathbb{R}^{d \times d} \), \( \mathbf{b}_G \in \mathbb{R}^d \) are learnable parameters, and \( \sigma \) is the sigmoid activation function. The final AVI representation \( \mathbf{H}_{GI} \) is obtained through element-wise multiplication \( \odot \) of \( \mathbf{g}_I \) and \( \mathbf{h}_{AS \to I} \):
\begin{equation}
    \mathbf{H}_{GI} = \mathbf{g}_I \odot \mathbf{h}_{AS \to I}
\end{equation}
\subsection{DualKanbaFormer Module}
The DualKanbaFormer comprises two specialized KanbaFormer modules: a Textual KanbaFormer for processing textual data and a Visual KanbaFormer for processing visual data. Each module is tailored to extract aspect-specific features from its respective modality, enabling effective multimodal sentiment analysis in MABSA.
\subsubsection{Textual KanbaFormer}
\paragraph{Textual KanFormer}
The Textual KanFormer processes textual data by taking the output of the ATI, denoted \( H_{AS} \), as its input. This module enhances the representation of aspect-related textual semantics using hierarchical sparse attention (ADSA), KANs and DyT. ADSA employs an aspect scope, aspect focus, and an aspect proximity to efficiently refine \( H_{AS} \), balancing global and local context while reducing computational complexity compared to dense MHA. The output is further processed through DyT normalizer to enhance inference stability and KANs to capture complex non-linear patterns, yielding a robust textual representation. In the Textual KanFormer, the input feature \( H_{AS} \in \mathbb{R}^{ts \times d} \) (where \( ts \) is the sequence length and \( d \) is the feature dimension) is transformed into query, key, and value representations for KanFormer. The query, key, and value representations are computed as follows: $q_{AS_{ts}} = W_{sq} H_{AS}$, $K_{AS_{ts}} = W_{sk} H_{AS}$, and $V_{AS_{ts}} = W_{sv} H_{AS}$, where \( W_{sq}, W_{sk}, W_{sv}\) are trainable projection matrices. For each branch $C \in \{\text{scp}, \text{fcs}, \text{prx}\}$, attention is computed across h heads.

\textbf{Aspect Scope}:
This branch captures the broad scope of the aspect across the entire sentence by compressing tokens \( K_{AS_{ts}} \) and \( V_{AS_{ts}} \) into coarse-grained representations. It reflects MABSA’s need to understand the extensive sentiment context tied to an aspect (e.g., general tone of a review).
\begin{equation}
    \tilde{K}_{AS_{ts}}^{scp} = \left\{ \phi(K_{AS_{id+1:id+l}}) \mid 1 \leq i \leq \left\lfloor \frac{ts - l}{d} \right\rfloor \right\}
\end{equation}
\begin{equation}
    \tilde{V}_{AS_{ts}}^{scp} = \left\{ \phi(V_{AS_{id+1:id+l}}) \mid 1 \leq i \leq \left\lfloor \frac{ts - l}{d} \right\rfloor \right\}
\end{equation}
\begin{equation}
    P^{\text{scp}}_{AS_{ts}} = \text{Softmax} \left( q_{AS_{ts}} (\tilde{K}_{AS_{ts}}^{\text{scp}})^T \right)
\end{equation}

where \( l \) is the block length, \( d \) is the stride, and \( \phi \) is a learnable MLP with intra-block position encoding. This reduces the sequence length \( ts \) to a smaller set of global summaries \( \lfloor ts / l \rfloor \), essential for modeling broad semantic relationships.

\textbf{Aspect Focus}:
This branch focuses on fine-grained, aspect specific textual blocks, selecting the most relevant segments based on importance scores. It aligns with MABSA’s goal of pinpointing sentiment bearing details directly linked to the aspect. To streamline the process, we adopt a uniform blocking scheme across both scope and focus branches, setting 
\( l' = l = d \). As a result, the focus importance scores \(P^{\text{fcs}}_{AS_{ts}}\) are directly obtained from the scope branch scores, i.e., \( P^{\text{fcs}}_{AS_{ts}} = P^{\text{scp}}_{AS_{ts}} \).

\begin{equation}
    I_{ts} = \{ \text{top-} n (P_{AS_{ts}}^{fcs}) \}
\end{equation}
\begin{equation}
    \tilde{K}_{AS_{ts}}^{fcs} = \text{Cat}[K_{AS_{il'+1:(i+1)l'}} \mid i \in I_{ts}]
\end{equation}
\begin{equation}
    \tilde{V}_{AS_{ts}}^{fcs} = \text{Cat}[V_{AS_{il'+1:(i+1)l'}} \mid i \in I_{ts}]
\end{equation}

where \( n \) is the number of selected blocks. This branch ensures precise attention to key textual segments (e.g., sentiment-bearing phrases).

\textbf{Aspect Proximity}:
This branch emphasizes proximity by retaining a fixed window of recent tokens near the aspect, ensuring local sentiment context is preserved. It’s critical in MABSA for capturing nearby modifiers or descriptors that refine aspect level sentiment.

\begin{equation}
    \tilde{K}_{AS_{ts}}^{prx} = K_{AS_{\max(0, ts - w):ts}}
\end{equation}
\begin{equation}
    \tilde{V}_{AS_{ts}}^{prx} = V_{AS_{\max(0, ts - w):ts}}
\end{equation}
where \( w \) is the window size, crucial for maintaining short-range dependencies around the aspect position.

\textbf{Attention and Gating}:
The outputs are combined using learned gates:
\begin{equation}
    \tilde{H}_{AS_{ts}} = \sum_{c \in C} g_{ts}^c \cdot \text{Attn}(q_{AS_{ts}}, \tilde{K}_{AS_{ts}}^c, \tilde{V}_{AS_{ts}}^c)
\end{equation}
where \( g_{t_s}^c \in [0,1] \) is the gate score for branch \( c \), derived from input features via an MLP and sigmoid activation.
\begin{equation}
    \text{Attn}(q_{AS_{ts}}, \tilde{K}^{c}_{AS_{ts}}, \tilde{V}^{c}_{AS_{ts}}) =
\sum_{i=1}^{ts} \frac{\alpha_{ts, i} \tilde{v}^{c}_{i}}{\sum_{j=1}^{ts} \alpha_{ts, j}}, 
\end{equation}
\begin{equation}
    \quad \alpha_{ts, i} = e^{\frac{q_{AS_{ts}}^\top \tilde{k}^{c}_{i}}{d_k}}
\end{equation}
Here, \( \alpha_{ts,i} \) denotes the attention weight between \( q_{AS_{ts}} \) and \( k_i \), while \( d_k \) represents the feature dimension of the keys.

The output is refined with a residual connection and passed through a KAN layer:
\begin{equation}
    \tilde{H}_{tK_{ts}} = DyT(\tilde{H}_{AS_{ts}} + H_{AS_{ts}})
\end{equation}
\begin{equation}
    H_{tK_{ts}} = DyT(\text{KAN}(\tilde{H}_{AS_{ts}}) + \tilde{H}_{AS_{ts}})
\end{equation}
\begin{equation}
    {DyT}(x) = \gamma \cdot \tanh(\alpha x) + \beta
\end{equation}

\paragraph{Textual Mamba} Mamba’s selective mechanism models long-range dependencies between aspects and opinion words, ensuring the global sentiment context is preserved without processing irrelevant tokens uniformly. When the Textual Mamba module receives an input \( \mathbf{H}_{AS_{ts}} \), it expands the dimension using two linear projections. Mamba passes the extended embedding through a SiLU activation and a convolution for one projection before feeding it into the SSM. The discretized SSM module's core can pick out knowledge relevant to the aspect and filter out noise. Through a multiplicative gate, the output of the SSM module is coupled with the other projection that follows SiLU activation as a residual connection. Lastly, the output \( \mathbf{H}_{\text{tM}}\) is provided by Mamba using an output linear projection. The process is formulated as follows:

\begin{equation}
    \mathbf{H}_{\text{tM}} = \text{Mamba}(H_{AS})
\end{equation}

\paragraph{Gated Fusion Mechanism} To dynamically assimilate valuable insights from the Textual KanFormer and Textual Mamba, we employ a gated fusion mechanism to reduce interference from unrelated data. This module uses a simple addition-based fusion mechanism to achieve gating, which controls the flow of information through gate maps, as shown in Fig.~\ref{fig:DualKanbaFormer}. The process is highlighted below:
\begin{equation}
    \text{G}_{tK} = \sigma(\text{CNN}(\mathbf{H}_{tK}))
\end{equation}
\begin{equation}
    \text{G}_{\text{tM}} = \sigma(\text{CNN}(\mathbf{H}_{\text{tM}}))
\end{equation}
\begin{equation}
\mathbf{H}_{\text{T}} = G_{\text{tK}} \odot \mathbf{H}_{\text{tK}} + (1 - G_{\text{tK}}) \odot G_{\text{tM}} \odot \mathbf{H}_{\text{tM}}
\end{equation}

where \( \sigma \) represents the sigmoid activation function, and \( \odot \) denotes element-wise multiplication.

\subsubsection{Visual KanbaFormer}

\paragraph{Visual KanFormer} 
The Visual KanFormer processes visual data from the Aspect-Visual Interaction (AVI) output, $\mathbf{H}_{GI} \in \mathbb{R}^{ti \times d}$, where $ti$ is the visual sequence length and $d$ is the feature dimension. Like the Textual KanFormer, it employs ADSA’s hierarchical sparse strategy to refine $\mathbf{H}_{GI}$, effectively balancing global context and local details. KANs capture complex non-linear patterns, and the DyT module enhances inference stability, resulting in robust visual representations. The components mirror those used in the Textual KanFormer, ensuring consistent processing across modalities. Given the visual input representation $\mathbf{H}_{GI}$, we obtain the query, key, and value matrices as follows: $q_{GI_{ti}} = W_{iq} \mathbf{H}_{GI}$, $K_{GI_{ti}} = W_{ik} \mathbf{H}_{GI}$, and $V_{GI_{ti}} = W_{iv} \mathbf{H}_{GI}$, where $W_{iq}$, $W_{ik}$, and $W_{iv}$ are trainable projection matrices.

\textbf{Aspect Scope}:
This branch captures the comprehensive semantic information embedded in the visual data by aggregating the key and value tokens $\mathbf{K}_{GI}$ and $\mathbf{V}_{GI}$ into coarse-grained representations:
\begin{equation}
    \tilde{K}_{GI_{ti}}^{scp} = \left\{ \phi(K_{GI_{id+1:id+l}}) \mid 1 \leq i \leq \left\lfloor \frac{ti - l}{d} \right\rfloor \right\}
\end{equation}
\begin{equation}
    \tilde{V}_{GI_{ti}}^{scp} = \left\{ \phi(V_{GI_{id+1:id+l}}) \mid 1 \leq i \leq \left\lfloor \frac{ti - l}{d} \right\rfloor \right\}
\end{equation}
\begin{equation}
    P^{\text{scp}}_{GI_{ti}} = \text{Softmax} \left( q_{GI_{ti}} (\tilde{K}_{GI_{ti}}^{\text{scp}})^T \right)
\end{equation}

\textbf{Aspect Focus}:
This branch focuses on fine-grained, aspect-specific visual blocks, selecting the most relevant segments based on importance scores from the Scope branch. We adopt a uniform blocking scheme, setting $l' = l = d$, allowing reuse of the focus scores i.e , \( P^{\text{fcs}}_{GI_{ti}} = P^{\text{scp}}_{GI_{ti}} \).
\begin{equation}
    I_{ti} = \{ \text{top-} n (P_{GI_{ti}}^{fcs}) \}
\end{equation}
\begin{equation}
    \tilde{K}_{GI_{ti}}^{fcs} = \text{Cat}[K_{GI_{il'+1:(i+1)l'}} \mid i \in I_{ti}]
\end{equation}
\begin{equation}
    \tilde{V}_{GI_{ti}}^{fcs} = \text{Cat}[V_{GI_{il'+1:(i+1)l'}} \mid i \in I_{ti}]
\end{equation}

\textbf{Aspect Proximity}:
This branch retains a fixed window of recent visual tokens near the aspect, preserving short-range visual dependencies:

\begin{equation}
    \tilde{K}_{GI_{ti}}^{prx} = K_{GI_{\max(0, ti - w):ti}}
\end{equation}
\begin{equation}
    \tilde{V}_{GI_{ti}}^{prx} = V_{GI_{\max(0, ti - w):ti}}
\end{equation}

\textbf{Attention and Gating}:
The final ADSA output is the gated sum of the three branches:
\begin{equation}
    \tilde{H}_{GI_{ti}} = \sum_{c \in C} g_{ti}^c \cdot \text{Attn}(q_{GI_{ti}}, \tilde{K}_{GI_{ti}}^c, \tilde{V}_{GI_{ti}}^c)
\end{equation}

The ADSA output is refined with a residual connection and passed through a KAN layer:

\begin{equation}
    \tilde{H}_{vK_{ti}} = DyT(\tilde{H}_{GI_{ti}} + H_{GI_{ti}})
\end{equation}
\begin{equation}
    H_{vK_{ti}} = DyT(\text{KAN}(\tilde{H}_{GI_{ti}}) + \tilde{H}_{GI_{ti}})
\end{equation}

\paragraph{Visual Mamba} 
Mamba’s selective mechanism allows Visual KanbaFormer to prioritize salient image regions, capturing the global semantic richness efficiently. The Visual Mamba module operates similarly to the textual Mamba but uses \( \mathbf{H}_{GI} \) as its input. The process is described below:
\begin{equation}
    \mathbf{H}_{\text{vM}} = \text{Mamba}(\mathbf{H}_{GI})
\end{equation}

\paragraph{Gated Fusion Mechanism}
To dynamically assimilate valuable insights from the Visual KanFormer and Visual Mamba, we employ a gated fusion mechanism to reduce interference from unrelated data. This module uses a simple addition-based fusion mechanism to achieve gating, which controls the flow of information through gate maps, as shown in Fig.~\ref{fig:DualKanbaFormer}. The process is defined as follows:
\begin{equation}
    \text{G}_{\text{vK}} = \sigma(\text{CNN}(\mathbf{H}_{\text{vK}}))
\end{equation}
\begin{equation}
    \text{G}_{\text{vM}} = \sigma(\text{CNN}(\mathbf{H}_{\text{vM}}))
\end{equation}
\begin{equation}
    \mathbf{H}_{\text{V}} = \text{G}_{\text{vK}} \odot \mathbf{H}_{\text{vK}} + (1 - \text{G}_{\text{vK}}) \odot \text{G}_{\text{vM}} \odot \mathbf{H}_{\text{vM}}
\end{equation}

where \( \sigma \) represents the sigmoid activation function, and \( \odot \) denotes element-wise multiplication.
\subsection{Multimodal Gated Fusion Module}
To reconcile the differences between the visual feature \( \mathbf{H}_{\text{V}} \) and the textual feature \( \mathbf{H}_{\text{T}} \), and to establish their correlation, we employ a gating mechanism similar to that used in the textual and visual KanbaFormer modules. The process is formulated as follows:
\begin{equation}
    \text{G}_{\text{T}} = \sigma(\text{CNN}(\mathbf{H}_{\text{T}}))
\end{equation}
\begin{equation}
    \text{G}_{\text{V}} = \sigma(\text{CNN}(\mathbf{H}_{\text{V}}))
\end{equation}
\begin{equation}
    \mathbf{H}_{C} = \text{G}_{\text{V}} \odot \mathbf{H}_{\text{V}} + (1 - \text{G}_{\text{V}}) \odot \text{G}_{\text{T}} \odot \mathbf{H}_{\text{T}}
\end{equation}

We employ mean pooling to condense the contextualized embeddings \( \mathbf{H}_{C} \), aiding downstream classification tasks. Subsequently, a linear classifier is applied to generate logits. Finally, the softmax transformation converts these logits into probabilities, enabling MABSA. The process is described as:
\begin{equation}
    \mathbf{H}_{\text{mp}} = \text{MeanPooling}(\mathbf{H}_{C})
\end{equation}
\begin{equation}
    p(a) = \text{softmax}(\mathbf{W}_{p} \mathbf{H}_{\text{mp}} + \mathbf{b}_{p})
\end{equation}

where \( \mathbf{W}_{p} \) and \( \mathbf{b}_{p} \) are trainable parameters comprising learnable weights and biases.

\subsection{Training}

We utilize the standard cross-entropy loss as the primary objective function:
\begin{equation}
    L(\theta) = - \sum_{(s, v, a) \in \mathcal{D}} \sum_{c \in \mathcal{C}} \log p(a)
\end{equation}
The loss is computed over all sentences, images, and aspects in the dataset \( \mathcal{D} \). For each triplet \( (s, v, a) \), representing a sentence \( s \), an image \( v \), and an aspect \( a \), we compute the negative log-likelihood of the predicted sentiment polarity \( p(a) \). Here, \( \theta \) encompasses all trainable parameters, and \( \mathcal{C} \) denotes the set of sentiment polarities.

\section{Experiments}
\subsection{Datasets}

We conduct experiments on two publicly available multimodal datasets, Twitter-15 and Twitter-17 \cite{Yu2019}, which comprise user tweets made between 2014 and 2015, and between 2016 and 2017, respectively, to assess the impact of the DualKanbaFormer framework. Table~\ref{tab:dataset_statistics} provides general information for both datasets.

\begin{table}[h]
\centering
\caption{Statistics for the experimental datasets.}
\label{tab:dataset_statistics}
\begin{tabular}{lcccc}
\hline
\textbf{Dataset} & \textbf{Division} & \textbf{Pos} & \textbf{Neg} & \textbf{Neu} \\
\hline
Twitter-2015 & Train & 938  & 368  & 1883 \\
             & Dev   & 303  & 149  & 670  \\
             & Test  & 317  & 113  & 607  \\
\hline
Twitter-2017 & Train & 1508 & 416  & 1638 \\
             & Dev   & 515  & 144  & 517  \\
             & Test  & 493  & 168  & 573  \\
\hline
\end{tabular}
\end{table}
\subsection{Implementation Details}

The DualKanbaFormer model is implemented in PyTorch, using pre-trained RoBERTa (BERT\textsubscript{LARGE}, 24 layers, 355M parameters) for 768-dimensional textual features and ResNet-152 for visual features. The Mamba layer utilizes 2 convolutional filters and a 16-dimensional state vector. ADSA is configured with a sliding window size of 2, compress block size of 4, compress block sliding stride of 2, selection block size of 4, and 2 selected blocks. Input sentences are standardized to 128 tokens. Training employs the Adam optimizer \cite{Kingma2014} (learning rate $1 \times 10^{-5}$), for 10 epochs, using DyT normalizer \cite{zhu2025norm} for stability, early stopping after 5 stagnant epochs, a batch size of 32, and 0.5 dropout to prevent overfitting.
\begin{table}[ht]
\centering
\caption{Experimental results on two publicly available datasets (Twitter-15 and Twitter-17). The best-performing results for each dataset are highlighted in bold.}
\begin{tabular}{l|c|c|c|c}
\hline
\multirow{2}{*}{\textbf{Model}} & \multicolumn{2}{c|}{\textbf{Twitter-15}} & \multicolumn{2}{c}{\textbf{Twitter-17}} \\
\cline{2-5}
 & \textbf{Acc.} & \textbf{F1} & \textbf{Acc.} & \textbf{F1} \\
\hline
\textbf{Text} & & & & \\
\hline
AE-LSTM                & 70.30 & 63.43 & 61.67 & 57.97 \\
RAM                    & 70.68 & 63.05 & 64.42 & 61.01 \\
MGAN                   & 71.17 & 64.21 & 64.75 & 61.46 \\
BERT                   & 74.15 & 68.86 & 68.15 & 65.23 \\
AMIFN                  & \textbf{77.63} & 73.51 & 70.09 & 69.10 \\
\textbf{DualKanbaformer} & 76.86 & \textbf{73.63} & \textbf{70.87} & \textbf{69.76} \\
\hline
\textbf{Text+Visual} & & & & \\
\hline
MIMN                   & 71.84 & 65.69 & 66.37 & 63.04 \\
TomBERT                & 76.18 & 71.27 & 70.50 & 68.04 \\
ESAFN                  & 73.38 & 67.37 & 67.83 & 64.22 \\
KEF-TomBERT            & 78.68 & 73.75 & 72.12 & 69.96 \\
ITM                    & 78.27 & 75.28 & 71.64 & 69.58 \\
AMIFN                  & 78.69 & 75.50 & 72.29 & 70.21 \\
GLFFCA                 & 77.72 & 74.21 & 71.15 & 69.45 \\
\textbf{DualKanbaformer} & \textbf{78.87} & \textbf{75.67} & \textbf{72.82} & \textbf{71.51} \\
\hline
\end{tabular}
\label{tab:results}
\end{table}

\subsection{Compared systems}
We evaluate the performance of our proposed DualKanbaFormer model against both text-based \textbf{AE-LSTM} \cite{Wang2016}, \textbf{IAN} \cite{Ma2017}, \textbf{MGAN} \cite{Fan2018}, \textbf{BERT} \cite{Devlin2018}, \textbf{AMIFN} \cite{Yang2024b} and multimodal \textbf{MIMN} \cite{Xu2019}, \textbf{TomBERT} \cite{Yu2019}, \textbf{ESAFN} \cite{Yu2020}, \textbf{ITM} \cite{Yu2022}, \textbf{GLFFCA} \cite{Wang2024}, \cite{Yang2024c} methods below.

\subsection{Experimental Results}

Table~\ref{tab:results} summarizes the experimental results on two benchmark Twitter datasets, Twitter-2015 and Twitter-2017, for ABSA and MABSA tasks. Our proposed DualKanbaFormer consistently outperforms existing baselines across all evaluated MABSA tasks, demonstrating its robustness and effectiveness.

For text-only ABSA tasks, AE-LSTM exhibits the lowest performance due to its limited ability to capture aspect-specific contexts effectively. In contrast, DualKanbaFormer surpasses SOTA models, including AMIFN, achieving higher F1 score and higher accuracy on Twitter-2017. On Twitter-2015, DualKanbaFormer maintains superior performance, though it performs comparably to AMIFN in some cases, likely due to AMIFN’s strength in syntax-based processing. These results highlight DualKanbaFormer’s ability to capture global semantic richness in texts via Mamba, and capture precise aspect-focused representations with ADSA.

In multimodal settings, incorporating visual data significantly boosts performance over text-only baselines, underscoring the value of multimodal integration. DualKanbaFormer outperforms AMIFN and other SOTA models, achieving higher F1 score and higher accuracy on Twitter-2015 and Twitter-2017. This improvement stems from the Visual KanbaFormer’s use of ADSA for precise aspect-focused representations, complemented by Mamba to capture global semantic richness in images.

Moreover, images provide complementary sentiment clues that enrich the overall sentiment representation, further boosting analysis performance in multimodal tasks. However, models like MIMN and ESAFN exhibit lower performance due to the absence of pre-training mechanisms to better align textual and visual modalities.

\begin{table}[htbp]
\centering
\caption{Results of an ablation study on the two datasets.}
\begin{tabular}{lcc}
\toprule
\textbf{Model} & \textbf{Twitter-15} & \textbf{Twitter-17} \\
\midrule
DualKanbaFormer & 78.87 & 72.82 \\
w/o Mamba & 77.53 & 71.31 \\
w/o KanFormer & 76.66 & 69.87 \\
w/o KANs & 78.01 & 71.79 \\
w/o DyT & 77.92 & 71.55 \\
w/o Intra-modal gated fusion & 77.75 & 71.23 \\
w/o Multimodal gated fusion & 77.63 & 70.99 \\
\bottomrule
\end{tabular}
\label{tab:ablation_results}
\end{table}

\subsection{Ablation Study}
Table~\ref{tab:ablation_results} reports ablation experiments on the Twitter-2015 and Twitter-2017 datasets, evaluating DualKanbaFormer’s components for MABSA. Removing Mamba lowers accuracy by 1.34\% and 1.51\%, showing its role in capturing global semantics. Excluding the KanFormer block reduces performance by 2.21\% and 2.95\%, underscoring ADSA’s aspect-specific precision. Replacing KANs with a feedforward network (FFN) decreases F1 scores by 0.86\% and 1.03\%, confirming their non-linear expressivity. Substituting DyT with layer normalization results in a performance drop of 0.95\% and 1.27\%, highlighting DyT’s contribution to inference stability. Using an FFN instead of intra-modal gated fusion reduces performance by 1.12\% and 1.59\%, while replacing multimodal gated fusion lowers F1 scores by 1.24\% and 1.83\%. These results validate the synergy and necessity of all components in the DualKanbaFormer architecture.

\section{Conclusion}
In this work, we introduced DualKanbaFormer, a novel framework that advances MABSA through the seamless integration of parallel Textual KanbaFormer and Visual KanbaFormer modules. By synergistically combining Mamba’s selective state space modeling with ADSA, our model efficiently captures global semantic richness and long-range dependencies while prioritizing aspect-specific details across textual and visual modalities. The incorporation of KANs and DyT further enhances non-linear expressivity and inference stability, eliminating the need for extensive hyperparameter tuning. A novel multimodal gated fusion layer unifies intra-modal and cross-modal representations, ensuring precise and robust sentiment predictions. Experimental results on two publicly available datasets demonstrate the effectiveness of our strategy. 

\paragraph{Limitation}
The DualKanbaFormer framework, while effective for Multimodal Aspect-based Sentiment Analysis (MABSA), has a critical limitation. Its strength depends heavily on precise aspect selection to pinpoint sentiment-critical elements in text and images. Noisy or ambiguous inputs, common in Twitter-2015 and Twitter-2017 datasets, such as mislabeled aspects, colloquial expressions, or low-quality images, can cause incorrect aspect identification, significantly reducing sentiment analysis accuracy. For example, vague aspect terms like "vibes" or occluded visual features, such as blurred signage, impair the framework’s focus, compromising robustness.





\appendix

\section{Case Study}
\label{sec:appendix}

Table~\ref{tab:use_case} presents two illustrative examples analyzed by our DualKanbaFormer model on the Twitter-2015 and Twitter-2017 datasets for MABSA tasks, compared against the ITM baseline and an ablated DualKanbaFormer.
In the first example, the sentence "Future lost to Meek Mill in the competition" implies a negative sentiment for the aspect "Future" without directly stating an opinion. Both ITM and the ablated DualKanbaFormer fail to detect this implied sentiment, constrained by their reliance on local or linear patterns. In contrast, DualKanbaFormer accurately captures the sentiment. Mamba models global textual dependencies, connecting "Future" to the competition’s outcome. ADSA employs a hierarchical sparse strategy with three branches: Aspect Scope aggregates coarse-grained tokens to maintain broad context, Aspect Focus selects fine-grained, sentiment-critical tokens (e.g., "lost" tied to "Future"), and Aspect Proximity emphasizes local interactions around the aspect term (e.g., nearby modifiers). KANs capture complex, non-linear relationships, such as the conditional sentiment implied by the loss, while DyT stabilizes inference through adaptive feature scaling, enabling precise negative sentiment prediction.

\begin{table*}[t]
\centering
\caption{Prediction of DualKanbaFormer, ablated DualKanbaFormer (w/o DualKanbaFormer modules), and ITM on four samples. \checkmark\ and \texttimes\ denote correct and incorrect predictions, respectively.}
\begin{tabular}{@{}p{0.16\textwidth}p{0.34\textwidth}p{0.1\textwidth}p{0.3\textwidth}@{}}
\toprule
\textbf{Image} & \textbf{Text} & \textbf{Ground Truth} & \textbf{Model Predictions} \\ 
\midrule

\raisebox{-\totalheight}{\includegraphics[width=0.13\textwidth]{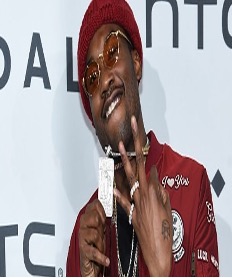}} &
(a) Meek Mill beats Drake, Kendrick Lamar, [Future]$_{neg}$ to win Top Rap Album at Billboard Music Awards &
NEG &
\scriptsize ITM: POS \texttimes \newline Ablated: POS \texttimes \newline DualKanbaFormer: NEG \checkmark \\ 
\midrule

\raisebox{-\totalheight}{\includegraphics[width=0.13\textwidth]{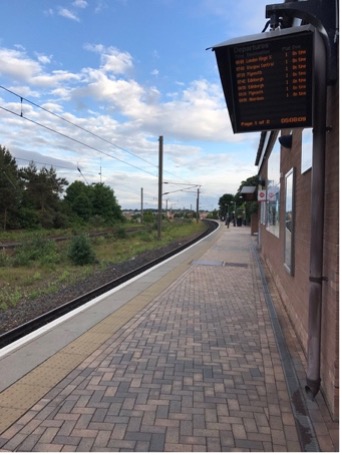}} &
(b) Off to [London]$_{pos}$ today to deliver training and meet clients. Nice quiet start to the morning in Dunbar station &
POS &
\scriptsize ITM: NEG \texttimes \newline Ablated: NEG \texttimes \newline DualKanbaFormer: POS \checkmark \\ 
\midrule

\raisebox{-\totalheight}{\includegraphics[width=0.13\textwidth]{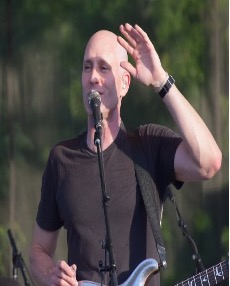}} &
(c) I loved seeing @ [VerticalHorizon]$_{pos}$ on Saturday at @ CelebrateFFX. Brilliant as expected. \#music \#summer \#concert &
POS &
\scriptsize ITM: POS \checkmark \newline Ablated: POS \checkmark \newline DualKanbaFormer: POS \checkmark \\ 
\midrule

\raisebox{-\totalheight}{\includegraphics[width=0.13\textwidth]{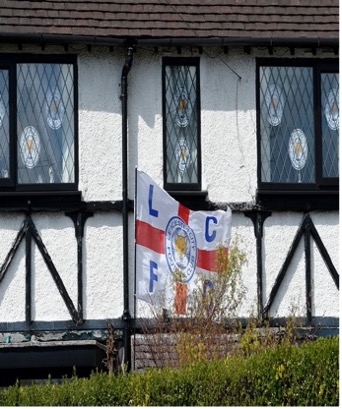}} &
(d) The final chapter of the fairytale – Leicester gear up for historic [Premier League]$_{neu}$ title … &
NEU &
\scriptsize ITM: NEU \checkmark \newline Ablated: NEU \checkmark \newline DualKanbaFormer: NEU \checkmark \\ 

\bottomrule
\end{tabular}
\label{tab:use_case}
\end{table*}

In the second example, a tweet about a "trip to London" is paired with an image of a train station featuring tracks, a platform, and clear signage, suggesting a positive sentiment through an orderly, calm atmosphere. ITM struggles with spatially dispersed visual cues, and the ablated model fails to integrate global context. DualKanbaFormer excels by leveraging Visual Mamba to capture global semantic richness, dynamically focusing on elements like signage and platform layout across the image. ADSA refines this with a hierarchical sparse approach: Aspect Scope aggregates broad visual regions to preserve scene context, Aspect Focus isolates sentiment-relevant patches (e.g., clear signage indicating order), and Aspect Proximity prioritizes local visual details near aspect-related elements (e.g., platform cleanliness). KANs model intricate interactions, such as signage clarity reinforcing positivity, and DyT ensures robust inference by stabilizing feature transformations, allowing DualKanbaFormer to correctly predict the positive sentiment.

\section{Effects of DualKanbaFormer Layers}
\label{sec:appendix}
\begin{figure}[!t]
\centering
\includegraphics[width=2.5in]{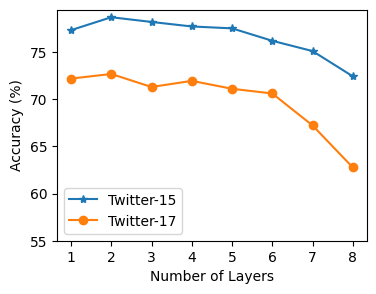}
\caption{Effect of different numbers of DualKanbaFormer layers}
\label{fig:kanbaformer_layers}
\end{figure}

According to our analysis, shown in Fig. \ref{fig:kanbaformer_layers}, the two layers produced the best results using the Twitter-15 and Twitter-17 datasets. Information about dependencies won't be sufficiently communicated if there are too few levels. Performance is reduced when a model has too many layers because it becomes overfit, and redundant data gets through. Many trials must be conducted to choose an appropriate layer number.
\section{DualKanbaFormer Efficiency Analysis}
\label{sec:appendix}
DualKanbaFormer optimizes computational efficiency in MABSA by leveraging Mamba, ADSA, KANs, and DyT. Mamba’s Selective State Space Model reduces complexity to near-linear $\mathcal{O}(t \log t)$, efficiently handling long textual and visual sequences compared to quadratic attention mechanisms. ADSA employs a sparse strategy—Aspect Scope, Focus, and Proximity—lowering complexity to $\mathcal{O}(t \cdot N_{\text{sparse}})$ by prioritizing aspect-relevant tokens and image patches. KANs utilize compact, learnable functions to model non-linear relationships with fewer parameters than standard feed-forward networks, thereby minimizing memory usage and inference costs. DyT replaces layer normalization with a lightweight, adaptive function, enhancing training efficiency. Together, these components enable DualKanbaFormer to process multimodal data with reduced resource demands. The model outperforms several SOTA on the Twitter-2015 and Twitter-2017 datasets, making it well-suited for scalable, real-time sentiment analysis applications.

\end{document}